# Intelligent prospector v2.0: exploration drill planning under epistemic model uncertainty

John Mern[1,3], Anthony Corso[3], Damian Burch[1], Kurt House[1] and Jef Caers[2]

[1]Kobold Metals, USA

[2]Stanford University, Department of Earth & Planetary Sciences, USA

[3]Terra Ai, USA

*Correspondence to*: Jef Caers (jcaers@stanford.edu)

**Abstract.**

Optimal Bayesian decision making on what geoscientific data to acquire requires stating a prior model of uncertainty. Data acquisition is then optimized by reducing uncertainty on some property of interest maximally, and on average. In the context of exploration, very few, sometimes no data at all, is available prior to data acquisition planning. The prior model therefore needs to include human interpretations on the nature of spatial variability, or on analogue data deemed relevant for the area being explored. In mineral exploration, for example, humans may rely on conceptual models on the genesis of the mineralization to define multiple hypotheses, each representing a specific spatial variability of mineralization. More often than not, after the data is acquired, all of the stated hypotheses may be proven incorrect, i.e. falsified, hence prior hypotheses need to be revised, or additional hypotheses generated. Planning data acquisition under wrong geological priors is likely to be inefficient since the estimated uncertainty on the target property is incorrect, hence uncertainty may not be reduced at all. In this paper, we develop an intelligent agent based on partially observable Markov decision processes that plans optimally in the case of multiple geological or geoscientific hypotheses on the nature of spatial variability. Additionally, the artificial intelligence is equipped with a method that allows detecting, early on, whether the human stated hypotheses are incorrect, thereby saving considerable expense in data acquisition. Our approach is tested on a sediment-hosted copper deposit, and the algorithm presented has aided in the characterization of an ultra high-grade deposit in Zambia in 2023.

## 1. Introduction

In the long term, the energy transition will demand a significant number of discoveries of critical minerals such as Copper, Nickel, Lithium Cobalt, and rare Earth elements. Exploration success in the discovery of large deposits of critical minerals and raw materials has been declining over the last three decades (Schodde, R.C., 2017), hence the average cost of making such discovery has increased significantly. At the current rate, the needed mineral resources will not be discovered in time. An important reason for this decline is that most major discoveries that are outcropping at the surface have been discovered (Davies et al., 2021; Savacool et al., 2020; Schodde, R.C., 2014). Exploration therefore needs to focus on a much harder problem: finding mineral resources under cover, without visible surface outcropping. This means that the role of geophysical & geochemical exploration and drilling will increase in importance. Geophysics, while essential in many exploration ventures, is not always successful in detecting ore bodies, in particular in discovering deeper, smaller, but larger grade resources. This is certainly true in the Copperbelt where high copper grades (over 5%) are contained in relatively thin layers termed ore shale located up to 1500m deep. Contrast this with very large porphyry systems at 0.5% grade mined with open pits. High grade deposits have the additional advantage of being much less destructive and use less water since they can be mined underground, as opposed to large scale open pit mining of many low-grade porphyry copper deposits, leading to significant environmental justice concerns (Agusdinata et al., 2018).

Artificial intelligence has recently received increasing attention to speed up discoveries. Before we proceed with the specific contributions of this paper, a review on what artificial intelligence is and how it is currently used in critical mineral exploration is necessary to avoid any confusion on the AI used in this work. First, unsupervised learning is a form of AI that searches the data for interesting patterns or anomalies in possibly large datasets, without having any reference labels such as known discoveries. In the mineral exploration context this data consists of geological, geochemical, and geophysical data from airborne, surface, or drilling (Wood, D., 2019). Orebodies are geochemical and mineralogical anomalies in the Earth, and often their composition remains not always well understood. Unsupervised learning can find both spatial, geophysical and compositional anomalies (e.g. Clare & Cohen, 2001; Abedi et al., 2013; Gazley et al., 2015; Zhang et al., 2019; Shirmard et al., 2022; Zuo & Carranza, 2023). For orebodies to be economical the need to be spatially compact, and they need to be compositionally distinguishable from the host-rock. Second, when label data (examples of true deposits) are available (such as in brown field exploration) meaning now some labelled data (truths) are available to train a machine learning model (Zuo, 2017; Dumakor-Dupey, & Arya, S., 2021; Jung & Choi, 2021). Advances in these topics have seen dramatic strides in recent years due to the advent of deep learning with neural networks. In mineral exploration, based on known deposits in some region of interest, mineral potential or mineral prospectivity maps provide insight into what are potentially fertile areas. It is however poorly documented how much these approaches have led to discoveries or failed to make discoveries by potential map guided drilling.

Eventually, discoveries are made using drilling, a time-consuming and expensive data acquisition method, requiring high accuracy targeting uncertainty (Pilger et al., 2001; Koppe et al., 2011; Koppe et al., 2017; Caers et al., 2022; Hall et al., 2022; Eidsvik, & Ellefmo, 2013; Soltani-Mohammadi & Hezarkhani, 2013; Bickel et al., 2018). In previous work, we proposed the use of Intelligent Agents, a form of AI that can perform sequential planning under uncertainty (Kochenderfer et al., 2022; Russell and Norvig, 2020), to aid with drill planning. This type of AI is used in self-driving cars, air traffic collision avoidance and playing games like chess (Brechtel et al, 2014; Grigorescu, 2020). The goal of this AI, which we termed "Intelligent Prospector," (Mern & Caers, 2023) has been shown to improve on the usual grid-spaced drilling, hence accelerate discovery and appraisal, reducing cost. The data used to train intelligent agents is very different from the data required for unsupervised and supervised learning. We will see how intelligent agents thrive on human intelligence about how the ore deposit was generated. Input to intelligent agents are human generated hypotheses as well as specific model assumptions on the nature of the 3D spatial variability of thickness (volume) or grade. The first version of the intelligent prospector was however limited in terms of the type of uncertainty one is dealing with. IPv1.0, like many similar approaches (Shahriari et al., 2016; Marchant et al. 2014) assumed that the major sources of uncertainty, such as epistemic uncertainty, are resolved and that spatial uncertainty, modelled though geostatistical methods, is the only form of uncertainty left, namely aleatoric uncertainty. Epistemic uncertainty, however, is critical in areas with few data, which is the norm in mineral discovery.

In February of 2024, the Financial Times (Dempsey, H., 2024) reported Kobold Metals had discovered an ultra-high-grade copper deposit in the Copperbelt of Zambia. Perhaps most remarkable about this news is that this is the first world-class body whose characterization was aided by artificial intelligence guided drilling, possibly opening a new avenue in an industry that has not yet fully embraced AI. The key contributing factor here is an intelligent agent that can plan for drilling includes now epistemic uncertainty. The aim of this paper is to present the artificial intelligence that helped to plan the characterisation of this ultra high-grade deposits (5% or more).

In the Copperbelt direct observation of the sediment-hosted copper is not always possible with geophysical or surface geochemical data. A second major challenge is that due the sparsity of data (sometimes no data at all), significant uncertainty exists on the nature, style and extent of the mineralization. Because of the lack of direct observation, geologists use a mineral systems approach (McCuaig & Hornsky, 2018) to speculate on the factors controlling such mineralization, often summarized in terms of a hypothesis, or hypotheses. For example, in the Copperbelt, the existence of faults, creating graben structures in an extensional rifting environment, in combination with the existence of a pyrite-bearing shale are critical components to create economical thickness and grade (Selly et al., 2018). A hypothesis is a statement of assumed truth, for example, about the presence of normal faults within the prospecting area. The opposite hypothesis is then the absence of such faults. Note that this statement is high level: it does not necessarily specify length, orientation and or location of these faults. Most times we do not directly observe the hypothesis in data; instead, it needs to be inferred from data, through inversion, meaning resulting in uncertainty. The latter requires a Bayesian approach as will become evident in the methodology. This type of uncertainty is termed: epistemic uncertainty, while the uncertainty of location and length are aleatory uncertainty (stochastic variation within the hypothesis). However, there is no guarantee that hypotheses stated by expert geologists contain the actual truth. It may be that after significant drilling one finds that all hypotheses are proven incorrect, in other worlds: falsified (Tarantola, 2006; Scheidt et al., 2018). The latter is a significant problem, as drilling under incorrect geological hypotheses may prove to be very expensive, since decision making relying on a wrong uncertainty model is ineffective in reducing uncertainty.

In this paper, we outline a completely new methodology for exploration drilling, using AI, never used before in any practice, whose first application helped to characterize a resource with grades over 5% copper. Critical to this first-time application in a real setting is the sequential planning of exploration drilling under epistemic (hypothesis) uncertainty, including the case where all human generated hypotheses may eventually be proven false. First, we describe some of the leading hypotheses of sediment-hosted copper and how they can be represented with geological models, then we describe the development of a partially observable decision process which can account for cases where all human stated hypotheses are incorrect. Finally, we compare this approach, on a range of hypothesis scenarios, against grid-based drilling.

## 2. Epistemic uncertainty when exploring in the Copperbelt

The Copperbelt is known for containing sediment-hosted copper deposits that have now been studied (Hitzman et al., 2005) for well over 100 years, and where major discoveries have already been made. Key to the generation of these deposits is the presence of an oxidized, metal-rich source rock, the availability of salts to facilitate the transport of these metals in brines, the presence of a reducing unit that will ultimately host the mineralization, and geodynamic processes that can drive fluid flow between the source and the reducing unit. Explanation of geological processes often proceeds using conceptual diagrams describing the source, transport and eventual trap (mechanical or chemical) of the mineralizing fluids (Selley et al., 2018; McCuaig et al., 2018). Holistically, geologists generally agree on how this worked for sediment-hosted copper deposits, but such knowledge is not specific enough yet to be helpful in quantitative decision making based on actual 3D geological models. For example, geologists may suggest the presence of a graben structure, but there is very little to no information that is specific to that area about the geometric nature of the faults that define such graben. Similarly, the pathway of copper-bearing fluids is dependent on the unknown permeability of rocks surrounding the shale, hence the specific extent of a precipitation front is not known either.

Because of the ongoing appraisal and mine planning at the Mingomba discovery, we cannot disclose any application of the intelligent prospector in that specific case. Instead, we create an analogue virtual case that has all the complexities of the actual case, specifically focusing on the problem of epistemic uncertainty. Our methodology is also not specific to drilling and/or sediment-hosted copper. As will be shown, it has application in data acquisition under epistemic uncertainty, a problem hardly limited to mineral exploration. Even though the geological modelling will be limited to 2D thickness and grade, the methodology extends to any form of 3D geological modelling.

The geological model used in our study is a 2D map containing the thickness and grade of the mineral deposit at each $(x, y)$ location. The mineral thickness is sampled from a Gaussian process, but the parameters of the Gaussian model are dependent on the presence or absence of a graben structure (where the mean thickness is higher within the graben). Similarly, we model the grade of the mineralization using a Gaussian process with parameters that depend on the presence of various geochemical alterations. The total mineralization at each grid point is computed by multiplying thickness by grade. To assess the economic value of a deposit, we make assumptions on the price per ton of ore, the costs to mine and process that ore, and assume that only ore content above a target grade is economical to extract. See Figure 1 for an example of the geological model and Appendix for a detailed description of the parameters used in the model.

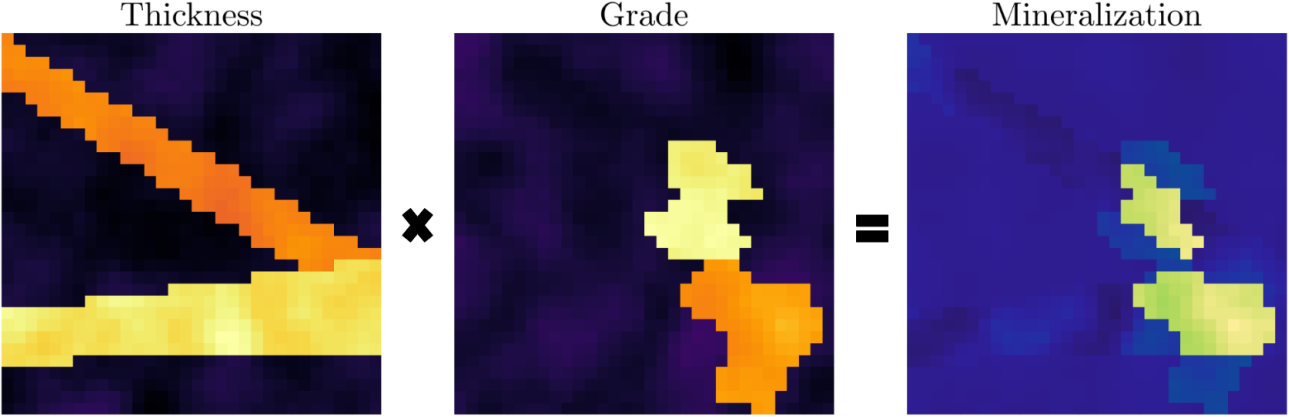

Figure 1: mineralization is a function of thickness controlled by graben as well as chemical alteration due to copper bearing fluids that precipitate in sedimentary layers

We define a class of non-overlapping hypotheses to serve as the conceptual geological models under study (see Figure 2). Each hypothesis is defined based on the number of graben structures and geochemical domains present in the geological model. Note that a hypothesis does not specify where exactly in the domain these grabens and geochemical structures occur.

The intelligent prospector will make decisions on where to drill boreholes. Each borehole provides an observation of both the grade and thickness of the mineralization at the borehole location but does not directly observe the hypothesis class that the geological model was sampled from. The graphical model that represents our modelling assumptions is shown in Figure 3. Although it is a simple model from a geological perspective it can be used to illustrate the challenges and opportunities of algorithmic decision making applied to problems with epistemic uncertainty.

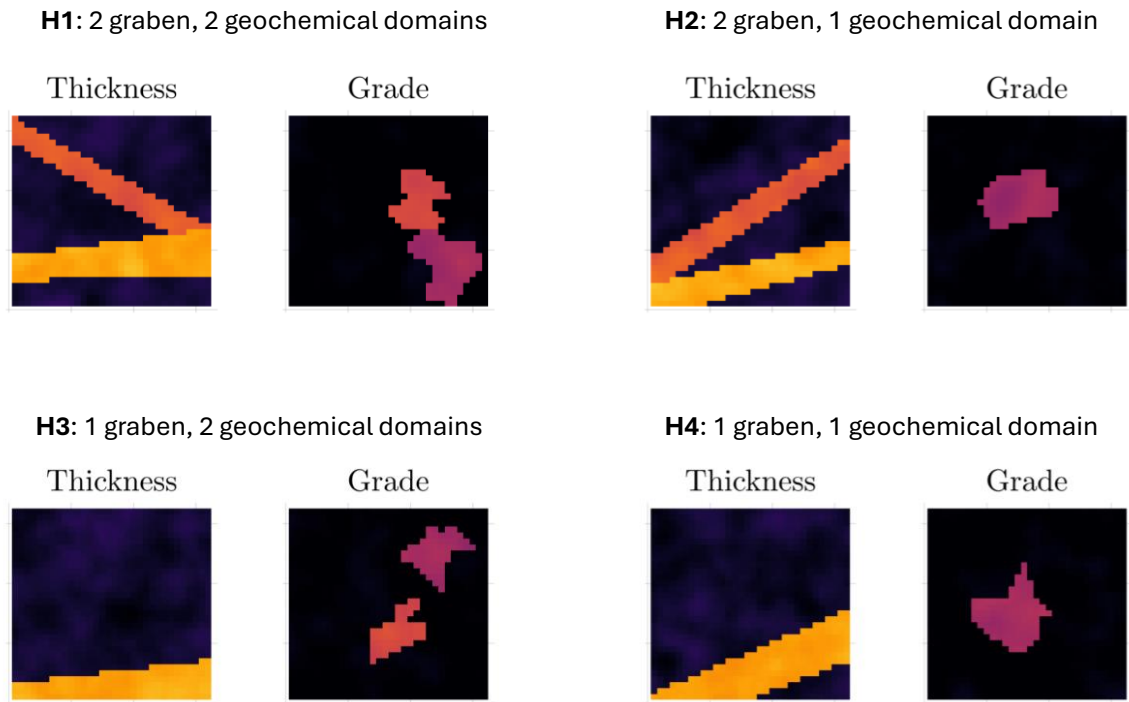

Figure 2: 4 Hypothesis Classes each with 1-2 Grabens, and 1-2 Geochemical domains where mineralization has been enhanced.

The objective of the case study is to assess the impact of algorithmic decision making on 1) the efficiency of exploration in the presence of aleatoric uncertainty (where the true model is part of one of the hypotheses), and 2) to develop an approach that can identify cases of high epistemic uncertainty (where the true model is not part of the hypotheses) and incorporate it into an algorithmic decision making framework. We conduct a series of experiments, where in each experiment, we select a ground truth model from a chosen hypothesis and then provide the intelligent prospector with 1 or more hypotheses to operate on (sometimes including the correct hypothesis and other times not). We then determine the efficacy of the intelligent prospector in terms of its accuracy and speed of reaching a final go/no-go decision and its ability to efficiently eliminate incorrect hypotheses.

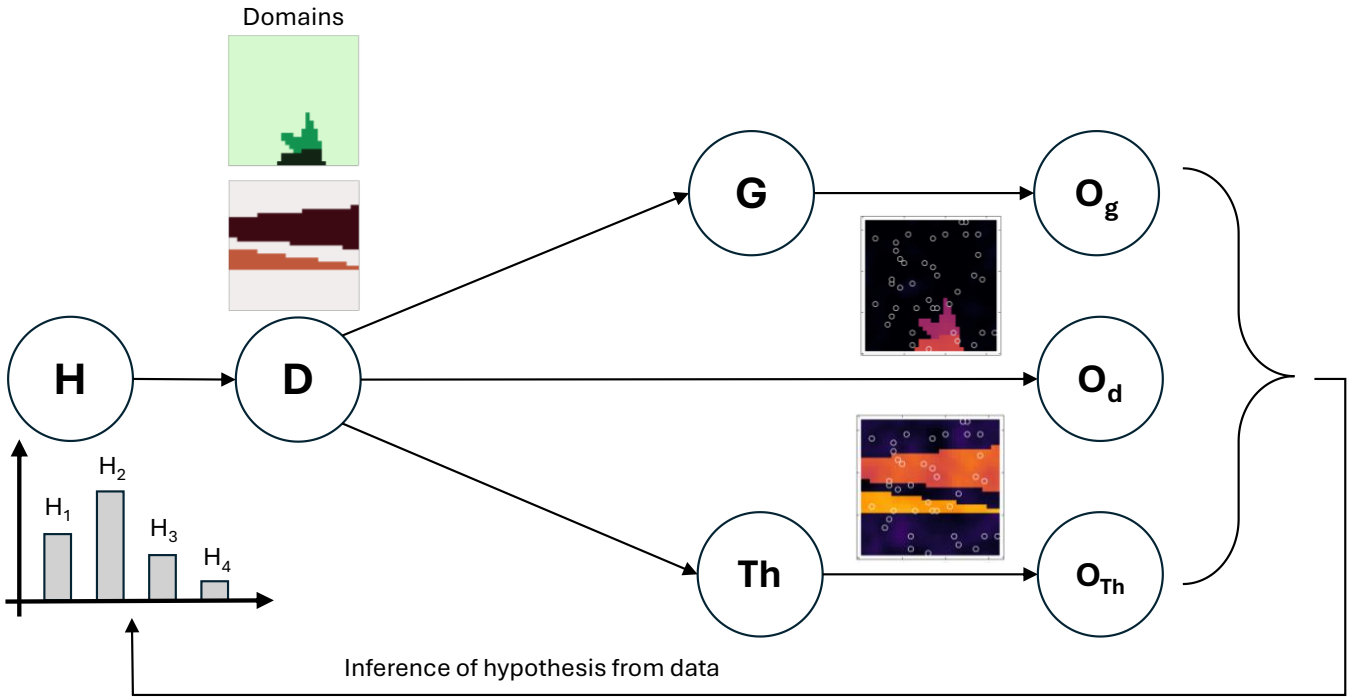

Figure 3: Bayesian network that expresses the relationship between the model variables, $H = hypothesis$, $D = Domain$, $G = grade$ and $Th = thickness$, borehole drilling observations ($O$) are on domain, grade and thickness at the drill location.

## 3. Methodology

### 3.1 Review of a POMDP formulation for sequential planning

In previous work (Mern & Caers, 2023) we framed any data acquisition campaign as a sequential decision process, which we will review again in this section. In a sequential problem, a decision-making agent must take a sequence of actions to reach an objective (goal or reward). Information gained from each action in the sequence can inform the choice of subsequent actions. An optimal action sequence will account for the expected information gain from each action and its impact on future decisions. This type of conditional planning may be referred to as closed-loop or feedback control. The actions considered in the paper are drilling locations or sequences thereof.

A sequential decision problem can be modelled formally as a Markov decision process (MDP). An MDP is a mathematical description of a sequential decision process defined by a collection of probability distributions, spaces, and functions. The full MDP is typically defined by the tuple $(S, A, T, r, \gamma)$. The state space $S$ is the space of all states that the decision-making problem may take at any step. In the mineral exploration process, the state is defined by the geological model of the subsurface deposit as well as the locations of existing drill holes. The action space $A$ defines the set of all actions that the agent may take. In mineral exploration drilling, this would be the set of all locations that the agent may possibly drill. The transition model $T(s_{t+1} \mid s_t, a_t)$, is the probability distribution over the next time step state $s_{t+1}$, conditioned on the current state and action. The step $t$ refers to the sequential actions and belief updates. The MDP formulation assumes that the state transition is fully informed by the immediately preceding state and action, which is the Markovian assumption. The transition model may be deterministic, as in the case of mineral exploration, where the underlying state doesn't change except for the location of additional bore holes, as they are drilled.

The reward function $r(s_t, a_t, s_{t+1}): S \times A \times S \to R$ gives a measure of how taking an action from a state contributes to the utility of the total action sequence which the agent seeks to maximize. The objective of an agent in an MDP is to maximize the sum of all rewards accumulated over an action sequence. To preference rewards earlier in the process, a time discount factor $\gamma \in (0,1]$ is used. The goal of solving an MDP is to maximize the sum of discounted rewards accumulated from a given state, defined as

$$\sum_{t=1}^{T} \gamma^{t-1} r(s_t, a_t, s_{t+1})$$

for a decision process with $T$ steps. The sum of discounted rewards expected from a state is defined as the value of the state $V(s)$. Given that the exact state transitions are not generally known in advance, the optimization target of solving an MDP is to maximize the expected value.

In many decision-making problems, and all real subsurface problems, the state at each time step (e.g. the 3D geological variability) is not fully known. In this case, agents make decisions based on imperfect observations of the relevant states of their environments. Sequential problems with state uncertainty are modelled as *partially observable* Markov decision processes (POMDPs). POMDPs are defined by the MDP tuple plus an observation space $O$ and a likelihood model for observations $L(o_{t+1} \mid s_{t+1}, a_t)$. The observation space defines all the observations that the agent may make after taking an action. Observations are generally noisy measurements of a subset of the state. The observation model defines the conditional distribution of the observation given the state and action.

To solve a POMDP, an agent must account for all the information gained from the sequence of previous observations when taking an action. It is common to represent the information gained from an observation sequence as a *belief*. A belief is a probability distribution over the unknown state of the world at a given time step. At the beginning of the decision-making process, the agent will start with a belief that is defined by all *prior* knowledge of the state available before making any observations. These include all model parameters that are uncertain and their prior distribution. Model parameters can be geological hypotheses, parameters within each hypothesis, and spatial variability (of e.g. grade) modelled on a grid. With each observation made, the belief is updated, typically using a Bayesian update as

$$b(s_{t+1}) \propto L(o_{t+1}|s_{t+1}, a_t)b(s_t).$$

Note that $b(s_{t+1})$ is AI notation for a posterior $p(s|o)$, where $p(s)$ is the prior. A belief may be an analytically defined probability distribution or an approximate distribution, such as a state ensemble updated with a particle filter.

Each decision in the sequence is made using the belief updated from the preceding observation. The process is depicted in Figure 4. An optimal choice in a sequential problem should consider all subsequent steps in the sequence. However, the number of trajectories of actions and observations reachable from a given state grows exponentially with the length of the sequence. As a result, optimizing conditional plans exactly is generally intractable. Instead, most POMDPs are solved approximately using stochastic planning and learning methods, which we cover in a later section.

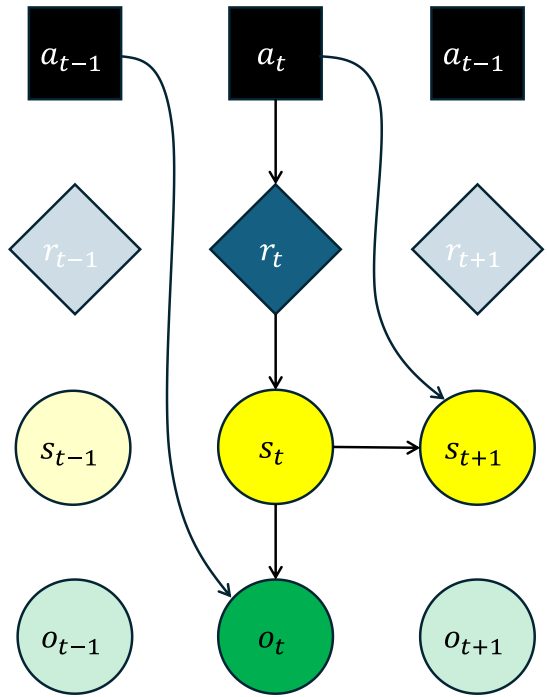

Figure 4: dynamic decision diagram of a partially observable Markov decision process.

**3.2 POMDP for exploration drilling under epistemic uncertainty**

**3.2.1 Overview**

Specifically, for our exploration drilling problem we have the following broad description of the POMDP model

1. **States:** joint state of geological state, hypothesis state, and drill states
    a. *Geological State:* spatial distribution of grade, thickness, and surface depth, defined over a cartesian grid.
    b. *Hypothesis State:* A tuple of the generating hypothesis scenario and a model of the domains over the exploration area. In a Bayesian context this is termed a *latent* variable (not directly observed)
    c. *Drill States:* The target pierce point and remaining time of each drill rig.
2. **Actions:** The locations that may be targeted for drilling, defined on a 2D cartesian grid. The set may be constrained by time-varying constraints.
3. **Observations:** grade and thickness extracted from drill-hole data

4. **Transitions:** The geological state is static. The drill states evolve deterministically according to a fixed drilling time and distance.
5. **Reward:** an economic model of revenue and costs

### 3.2.3 State space definition

Rewards depends on the model state, and if the full model state were observed, the reward would be perfectly known. This is not the case, and hence uncertainty in the state needs to be quantified. To do that we need to specify the random variables defining the geological model. Quite generally, one can define the random variables of a geological models through a hierarchy. At the highest level are model hypothesis (e.g. using a Gaussian process, an object model, a training image etc..), each of which may contain its own parametrization (e.g. the geometries of the domains contained in the model, the parameters of a variogram). Finally, using hypotheses and their parametrization, one can use geostatistical algorithms to generate model realizations in 3D space. Figure 3 shows the specific Bayesian network model that defines dependencies between model parameters and the data informing these parameters.

### 3.2.4 Belief model

Because the random variables defining the model are hierarchical, so will be any distribution (prior, likelihood & posterior). Prior distributions instantiate the lack of knowledge prior to any drilling, often using analogue information, while the posterior distribution (or belief) provides an update of the uncertainty on all model parameters. Important here is that grid model variables are directly observed (the grade and thickness in a drill hole), while others are indirectly observed (the hypothesis).

The belief over geological models involves two components: the model over the discrete domains and the model over the geological properties (in this case, grade and thickness). For the latter, Gaussian process models were used to model the spatial correlation of the properties, but the mean of the process was controlled by the local domain. The discrete domains are explicitly modelled based on a number of geometric parameters (described in more detail in the appendix) and implemented in Turing.jl. While belief updates on a Gaussian process can be performed analytically, the update of the parameters controlling the structural and geochemical domains are more challenging and involve maintaining distributions over the geometric parameters that describe the domain.

Specific to our case, the following hierarchical definition is used. The hypothesis is assumed to be a discrete choice (a categorical random variable) while parametrization and 3D models may contain a mix of categorical and continuous random variables. Following the Bayesian network of Figure 3, the prior distribution is stated as follows.

$$f(H, D_{chem}, D_{grab}, Th, g) = P(H)f(D_{chem}|H)f(D_{grab}|H)f(Th|D_{grab})f(g|D_{chem})$$

$H$ refers to the hypothesis, $D$ to domains, either "grab" is graben, or "chem" is geochemistry, $g$ to grade and $Th$ to thickness at each location. Sampling from this distribution follows the hierarchy of model variables. Further, we use the following, reasonable assumptions:

- Parameters between domains are independent, for example, grade and thickness distributions are spatially independent between domains.
- Hypothesis scenarios hold over the entire domain.

Explicitly stating the probability distributions (as opposed to algorithmically defined ones) allows calculating many important statistics:

- Calculating expected values and likelihoods
- Generating state samples
- Updating distributions given new data

**3.3.2 Belief updating**

To update distributions sequentially, we need the definition of a likelihood of the data accounting for the Bayesian hierarchy. The data variable $O = \{O_{chem}, O_{grab}, O_{Th}, O_G\}$ consists of observation in drill holes of domain, grade, thickness, hence we can decompose the likelihood distribution $\ell$ for given data acquisition action $a$ as follow

$$\ell(O|a) = \sum_{h \in H} \ell(O|h, a)P(h)$$

$$= \sum_{h \in H} P(h) \sum_{d_{chem}, d_{grab}} \ell(O|h, d_{chem}, d_{grab}, a)P(d_{chem}|h)P(d_{grab}|h)$$

$$= \sum_{h \in H} P(h) \sum_{d_{chem}, d_{grab}} \ell(O_{chem}|h, d_{chem}, a)\ell(O_{grab}|h, d_{grab}, a)\ell(O_{Th}|h, d, a)\ell(O_G|h, d, a)P(d_{grab}|h)$$

In this formulation, we rely on the hierarchical model definition in Figure 3. First, the law of total probability is used to define the marginal likelihood $\ell(O|a)$ in terms of the conditional distributions. The first sum is taken over each hypothesis index $h$. The second integral is taken over each possible geochemical and graben domains. Finally, one decomposes the likelihood of an observation into the likelihoods of each component (domains, grade, and thickness).

We make the additional assumption that likelihood of the grade and thickness observations are given by a normal distribution (Gaussian Process) whose mean $\mu$ and covariance $\sigma^2$ are obtained by kriging (Gaussian process regression), respectively. The domain observation is assumed deterministic, defined mathematically by the Kronecker delta distribution.

$$\ell(O_g|h, d_{chem}, a) = \mathcal{N}(O_g | \mu^a_{h, d_{chem}}, (\sigma^2)^a_{h, d_{chem}})$$

$$\ell(O_{Th}|h, d_{grab}, a) = \mathcal{N}(O_{Th} | \mu^a_{h, d_{grab}}, (\sigma^2)^a_{h, d_{grab}})$$

and for the domains

$$\ell(O_{chem}|h, d_{chem}, a) = \delta_{d_{chem}, a}(O_{D_{chem}})$$

$$\ell(O_{grab}|h, d_{grab}, a) = \delta_{d_{grab}, a}(O_{D_{grab}})$$

The likelihood of the grade and thickness observations are given by a normal distribution defined by kriging (hard) data. The domain observation is a discrete outcome, defined mathematically by the Kronecker delta distribution. We assume that the domain is directly observed, which may not always be the case. We can also calculate the uncertainty of mean and variance of grade and thickness given all uncertainties jointly.

For complex belief updates with non-analytical posterior distributions, it is common to apply sampling-based approximation techniques for the full posterior distributions. In this work we use Elliptical Slice Sampling (ESS, Murray et al., 2010) which is a Markov Chain Monte Carlo (MCMC) method used for sampling from posterior distributions that are approximated to be Gaussian or have a Gaussian prior. To apply it to this problem, we ensure that prior distributions over geometric parameters are Gaussian. ESS is particularly effective for high-dimensional distributions where traditional methods may struggle. The algorithm works by defining an elliptical contour around the current sample point, drawing a candidate point along this contour, and then accepting or rejecting the candidate based on its likelihood. This approach efficiently explores the posterior distribution, making it well-suited for scenarios where the prior distribution is Gaussian, and the likelihood is complex.

**3.2.5 Dealing with a falsified prior model**

It is certainly possible, if not the norm, that none of the hypotheses contain in the set $H$ are correct. What we mean by this, is that at some point, after drilling several boreholes the probability of each hypothesis outcome $P(H = h_i) = 0 \; \forall \; h_i$, the hypotheses are falsified (in the sense of Popper, see Caers, 2018). However, when probability distributions have infinite support (e.g. Gaussians, which are common in geostatistical modelling), the likelihood of observations (and therefore hypotheses) are never exactly 0, but instead get increasingly small. The challenge, however, is that a particular sequence of observations can have extremely small likelihoods, even when coming from the true hypothesis class, and it is difficult to distinguish this phenomenon from true falsification of the prior.

The way we approach this problem is by adding another hypothesis $h_0$ to the set $H$, termed a null hypothesis. This hypothesis is defined as a maximum entropy model, by assuming a high-entropy prior distributions over the observed variables $Th$ and $g$ and removing any spatial correlation. While there are many choices of non-spatially correlated, high entropy distributions, we select a mixture model of Gaussian distributions that produce similar histograms to the true underlying geology. In practice, the modelling choice can depend on knowledge of the range of property values in the prospect due to drilling. An example null hypothesis sample and corresponding histograms compared to one hypothesis is shown in Figure 5, and the model parameters are provided in the appendix.

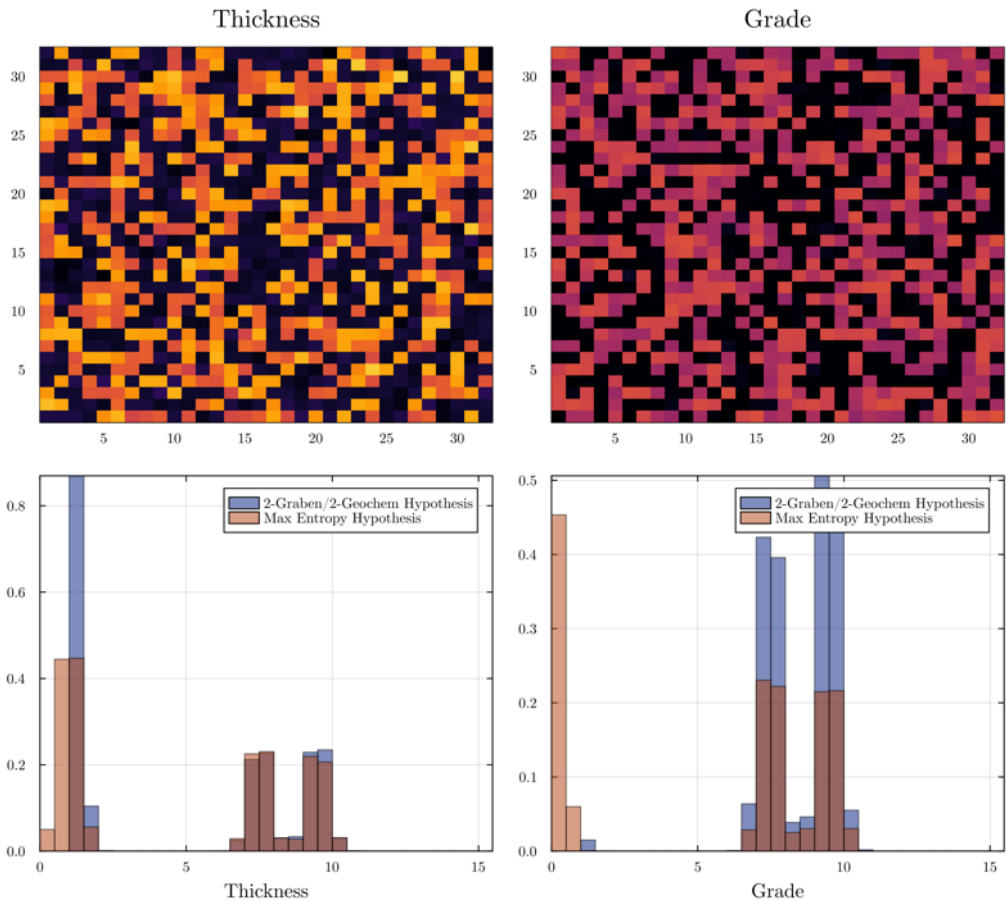

Figure 5: Null hypothesis example as a mixture of normal distributions. Normal distributions were selected to overlap with the grade and thickness values of the underlying geological model but contain no spatial structure.

The purpose here is to compare, during drilling, the likelihood of this null hypothesis with the likelihood of the other (human generated) hypotheses. If the likelihood of the null hypothesis becomes much larger than the human hypothesis, then this is an indication that humans are incorrect, because the random noise hypothesis is certainly incorrect ("geology is not random" is a very defensible statement). Figure 6 shows the evolution of the log-likelihood of several (incorrect and correct) hypotheses compared to the null hypothesis versus the number of bore holes that have been observed. We see initially, multiple hypotheses are plausible, but after just 2 bore holes, hypotheses 3 and 4 are falsified. Then after 7 bore holes hypothesis 2 is falsified leaving behind the correct hypothesis, hypothesis 1.

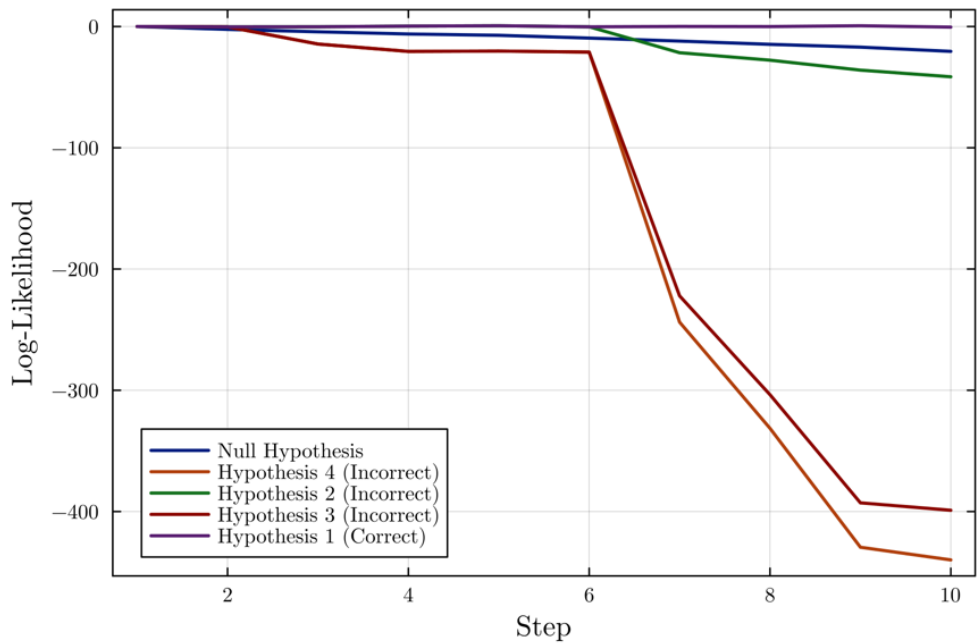

Figure 6: Log likelihood of various hypotheses vs. number of bore holes (=step). The correct hypothesis is never falsified while all incorrect hypotheses eventually are. After sufficient boreholes, the unrealistic null-hypothesis becomes more likely than any human specified hypothesis

When all prior hypotheses are falsified, additional models should be included into the belief of the intelligent prospector. There is currently no AI that can magically invent new hypotheses, this would still be the role of the human, at least for now. However, we can help the AI to detect this issue, thereby prompting humans to make more/better hypothesis, possibly with the insight of the data acquired by drilling.

**3.3 Solving the POMDP drill planner**

In the previous section, we formulated a mathematical model, based on a specific mineral exploration drilling problem which included various model assumptions. In this section, we present a method to solve the exploration drilling POMDP. Algorithms to solve POMDPs can typically be applied to any valid POMDP model, though with differing effectiveness. The remaining subsections are divided into the two primary tasks required to solve the POMDP: searching over the large, combinatorial space of possible action sequences and updating the belief with the new information obtained via drilling.

### 3.3.1 POMDP solvers

A common approach for solving POMDPs is to estimate an action-value function that computes the expected returns (or value) of taking a particular action in a particular belief:

$$\bar{Q}(b,a) = \frac{1}{n}\sum_{b'}\bar{V}(b')$$

The value function can be estimated via online and offline methods. Online methods are solved during execution, meaning each time an action is taken, a new plan is generated to find the next best action. Online methods often involve a form of forward search such as Monte Carlo tree search, where value function estimates are maintained during the search and there is an explicit trade-off between exploration and exploitation. Offline methods solve for the value function over all possible beliefs prior to execution. While this is intractable to perform exactly for even moderately sized problems, point-based methods where the possible beliefs are sampled, have shown strong performance. Point-based methods typically rely on discretised state, action and observation spaces in order to maintain a piecewise linear representation for the value function and the ability to perform exact belief updates. Online tree-search methods are scalable any-time algorithms that are easy to apply out of the box and have therefore previously been applied to a variety of subsurface problems (Wang et al., 2022, Wang et al. 2023). However, they suffer from high variance in their value function estimates and therefore rely on expert-defined heuristics and hyperparameter tuning to be effective. Conversely, offline methods have much lower variance and typically better performance compared to online methods but can be computationally more expensive.

In this work we opt for an offline POMDP algorithm for its superior consistency in performance. In particular, we use the Successive Approximations of the Reachable Set under Optimal Policies (SARSOP) algorithm (Kurniawati et al., 2009). SARSOP focuses on the subset of the belief space that is reachable under optimal policies, which significantly reduces computational complexity by concentrating on belief points that are most likely to be encountered during execution. This is achieved through a combination of forward and backward exploration that iteratively refines the belief space, allowing the algorithm to focus on high-probability regions while pruning less relevant areas. By doing so, SARSOP strikes a balance between computational efficiency and solution quality, making it particularly effective for high-dimensional and complex POMDPs.

To apply point-based methods like SARSOP to continuous POMDPs, it's necessary to discretize the continuous elements of the problem. The state space can be discretized by sampling from the belief distribution, allowing us to maintain a discrete ensemble of states that are representative of the underlying continuous space. This ensemble serves as the foundation for planning, with SARSOP operating over these sampled states. The action space is discretized by considering potential borehole locations on a regular grid over the domain, thereby transforming the continuous action space into a manageable set of discrete actions.

The most challenging aspect to discretize is the observation space. To tackle this, we generate a large number of observation samples by randomly selecting states and actions, and then use k-means clustering to group these observations into discrete clusters. Future continuous observations can then be converted into discrete form by assigning them to the nearest cluster. To determine the observation probability distribution, we sample all state-action pairs and calculate the frequency of each discrete observation. This approach allows us to convert the continuous POMDP into a discretized version that can be effectively solved using SARSOP. During execution, the belief update is performed on the continuous observation and the POMDP is then re-discretized and resolved via SARSOP to determine the next approximately optimal action. Finally, a detailed implementation of Intelligent Prospector v2.0 is publicly available (Corso, 2024).

## 4. Results

Maintaining multiple distinct geological hypotheses may enable decision makers to better understand the geological system during exploration and identify when none of the hypotheses are conforming to the observed data (through comparison with a null hypothesis). To characterize the effect of this modelling and decision-making approach we design experiments to answer two important questions. 1) will the increased aleatoric uncertainty of maintaining multiple geological hypotheses (one correct and multiple incorrect) degrade the performance of algorithmic decision-making approaches to drill targeting? 2) If the correct geological hypothesis is not represented in the belief, will an algorithmic decision-making approach falsify all incorrect hypotheses faster than baseline approaches?

In answering these questions, we use standard evaluation approaches for decision making systems (Kochenderfer et al, 2022). In all experiments, we select a ground truth geological model from the 2-graben, 2-geochemical domain hypothesis class and use that model to produce observations of the grade and thickness. For each experiment, the decision-making agent maintains a belief (different initial beliefs for different experiments) and recommends bore hole locations at each step. After an action is recommended the agent observes the noisy observation of the grade and thickness at the proposed bore hole location and re-plans for the next action. At each step the agent may decide to walk away from the project (concluding it is unlikely to be economical) or commit to developing the project (concluding that there is a high likelihood of profit). In each experiment we conduct this process with 17 different ground truth geological models and compute mean and variance statistics across those trials. We compare against a baseline approach of grid-drilling in a 6x6 pattern leading to a total of 36 bore holes, after which a go/no-go decision is made on a belief that has been updated with all 36 observations.

To address the first question, we seed the agent with 4 different hypotheses (those depicted in Figure 2), which includes the correct hypothesis. For each ground truth model, we record the error in the predicted ore value under the belief as a function of bore holes in order to determine how quickly the decision-making agent is able to estimate the deposit value compared to the baseline. We also record whether the correct go/no-go decision was reached by the agent and the baseline policy. We use a simple economic model with cut-off grade = 6, extraction cost = 35, and drill cost = 0.1. To address the second question, we remove the correct hypothesis class from the belief of the agent. First, we show that without the correct hypothesis in the belief, the agent may walk away from significantly economic deposits. Then using our null hypothesis comparison, we record the number of falsified hypotheses as a function of bore hole number for both the agent and the baseline policy. The experimental results are given below.

**4.1 Aleatoric Uncertainty**

The main point of this subsection is to show that having multiple hypotheses is a viable paradigm for representing a belief and performing drill targeting by solving the resulting POMDP. Figure 8 shows a significant improvement over a grid search baseline, in this case. For the same decision accuracy, the POMDP needed less than half as many boreholes as grid-based drilling. The drilling outcome for a specific example (one Monte Carlo simulated truth), comparing grid-based drilling (Figure 8) vs a POMDP planner (Figure 9) illustrated the significant difference in efficiency.

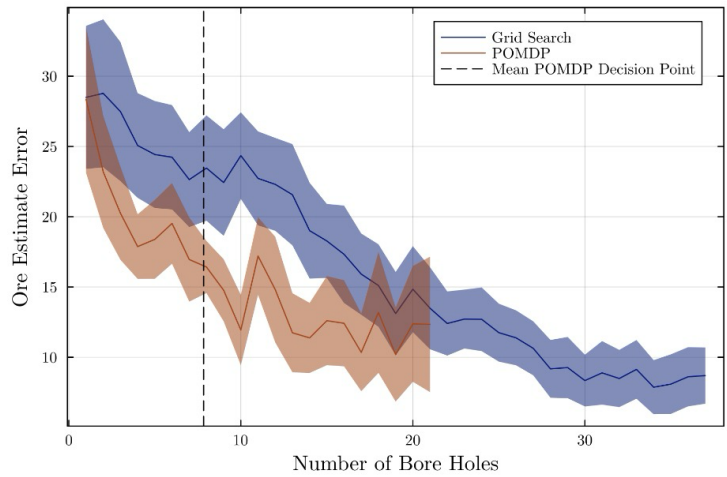

Figure 7: On the left, we plot the error in the estimated value of ore in the deposit as a function of the number of bore holes for a grid search baseline and the POMDP solution. On the right we present the go/no-go accuracy of the grid search and POMDP-based drill targeting approach and the mean number of bole holes required before a decision was made.

|  | Decision Accuracy | Number of Bore holes |
|---|---|---|
| Grid | 0.88 ± 0.08 | 36 |
| POMDP | 0.82 ± 0.09 | 16.06 ± 2.56 |

## After 2 boreholes

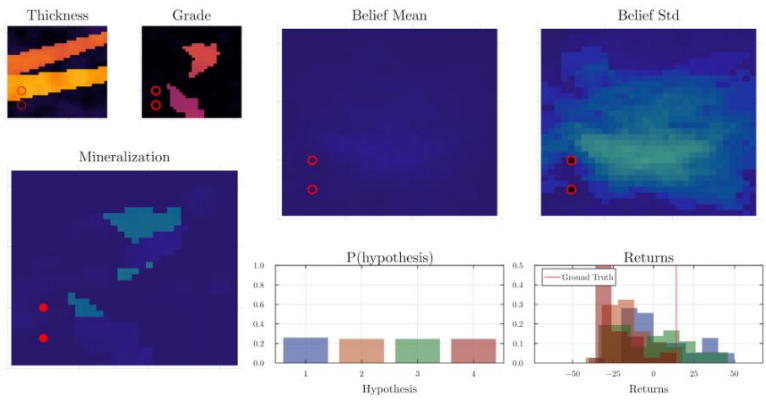

## After 7 boreholes

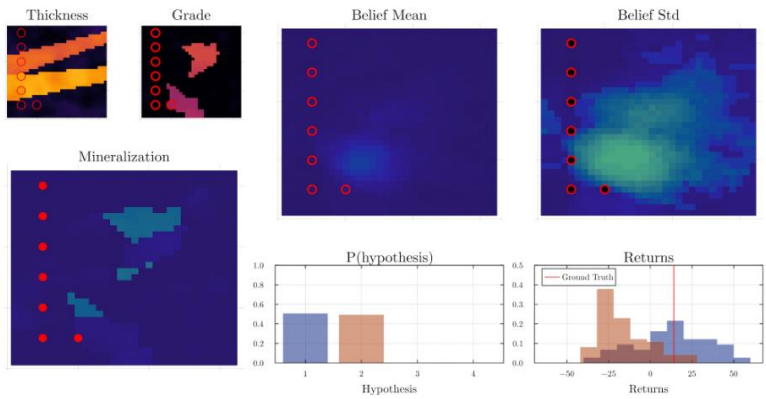

## After 35 boreholes

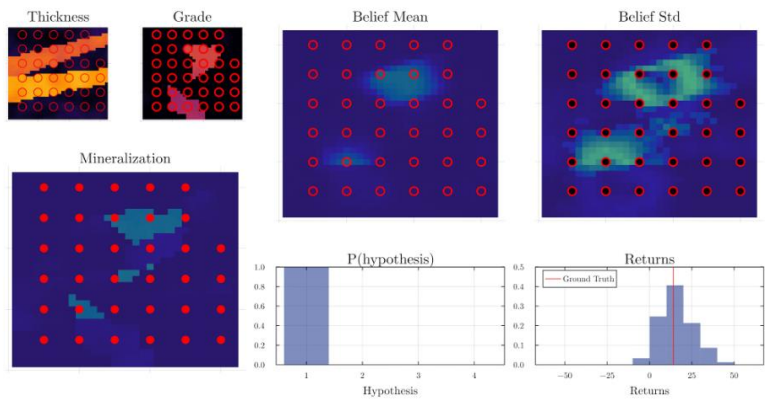

Figure 8: results of grid-drilling, for a few different amount of boreholes.

## After 2 boreholes

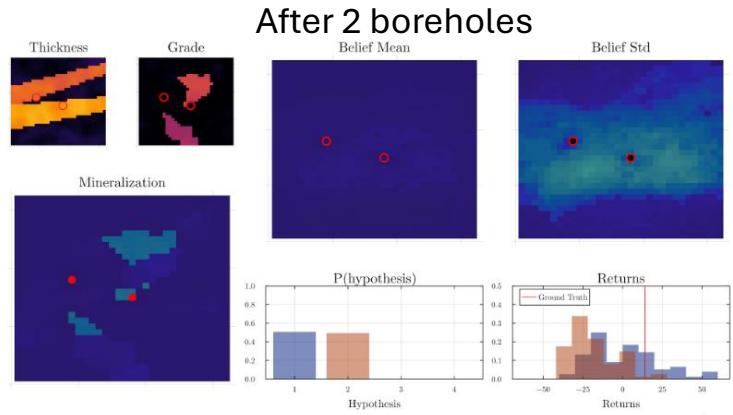

## After 7 boreholes

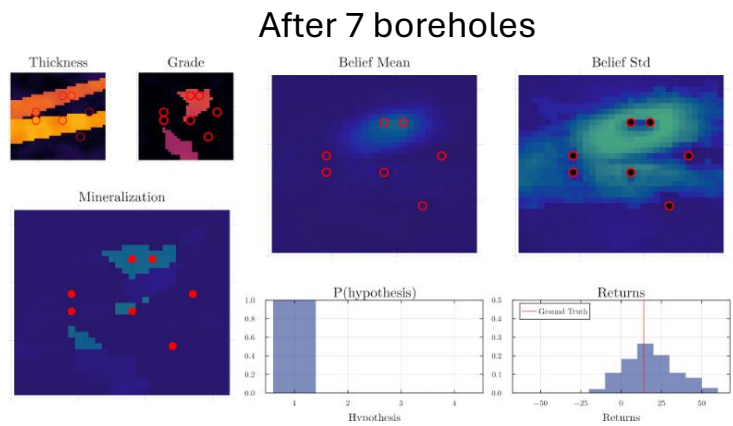

## Last (9th) borehole

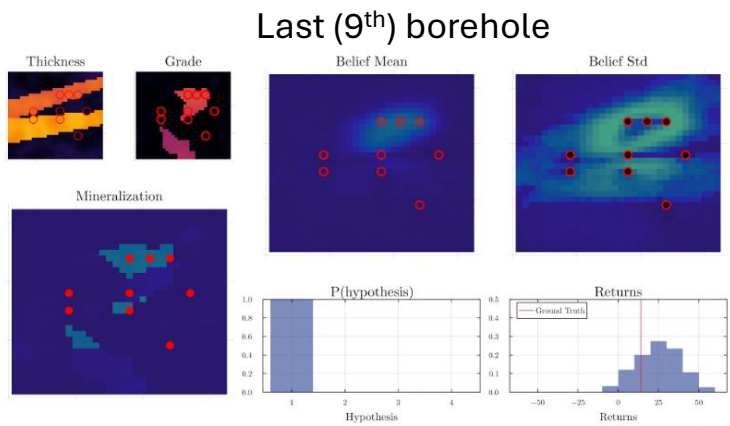

Figure 9: results of POMDP-based drilling, for a few different number of boreholes. Convergence is reached after only 9 boreholes

## 4.2 Drill planning under a falsified prior model hypothesis

Having shown that the decision-making agent can quickly discriminate between multiple incorrect hypotheses when the correct hypothesis class is present in the belief, we now explore the (realistic) setting where all hypotheses are incorrect. We first consider the case without any hypothesis falsification check and show a sample trajectory in Figure 10. Here we see that since the agent does not consider the presence of a second geochemical domain in its belief, it determines that the current ore is sub-economic and walks away from a prospect that is significantly profitable. This type of wrong decision is extremely costly and illustrates the need to have a wide range of hypotheses to maximize the chances of correctly modelling the ground truth geology.

Then we consider the problem of hypothesis falsification, by checking for falsified hypotheses (by comparing log likelihood of the data to the maximum entropy null hypothesis), after each new bore hole. We record the number of falsified bore holes (averaged over 17 trials) for both the POMDP agent and the grid baseline and report the results in Figure 11. We see that the bore holes selected by the POMDP agent are able to more quickly falsify the incorrect hypotheses (on average). We note that the POMDP agent is not explicitly optimizing for uncertainty reduction or falsification, but rather to determine the maximum expected reward. However, this result is evidence that reducing uncertainty and therefore falsifying hypotheses is a by-product of this simple profit-maximizing objective. Grid-drilling at this spatial resolution is not able to detect that all hypotheses are false.

## After 3 boreholes

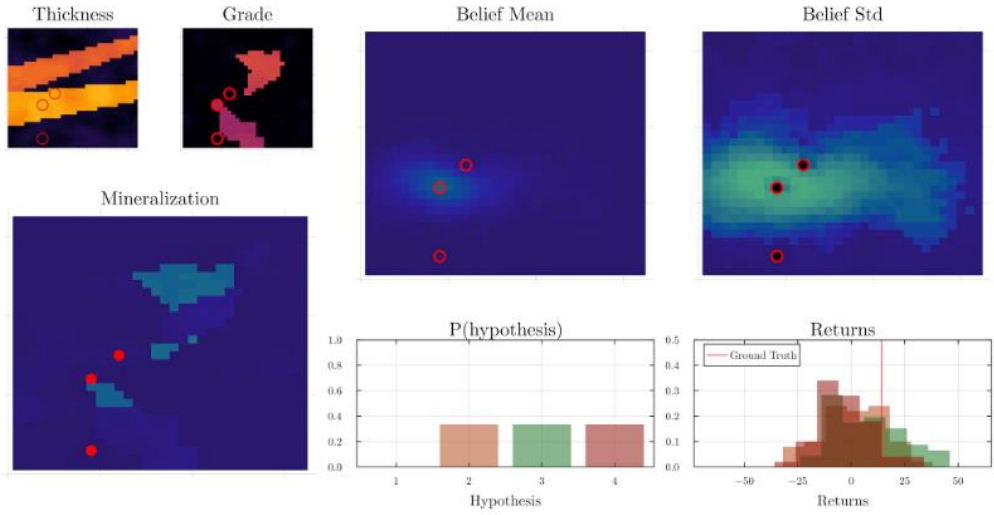

## After 8 boreholes

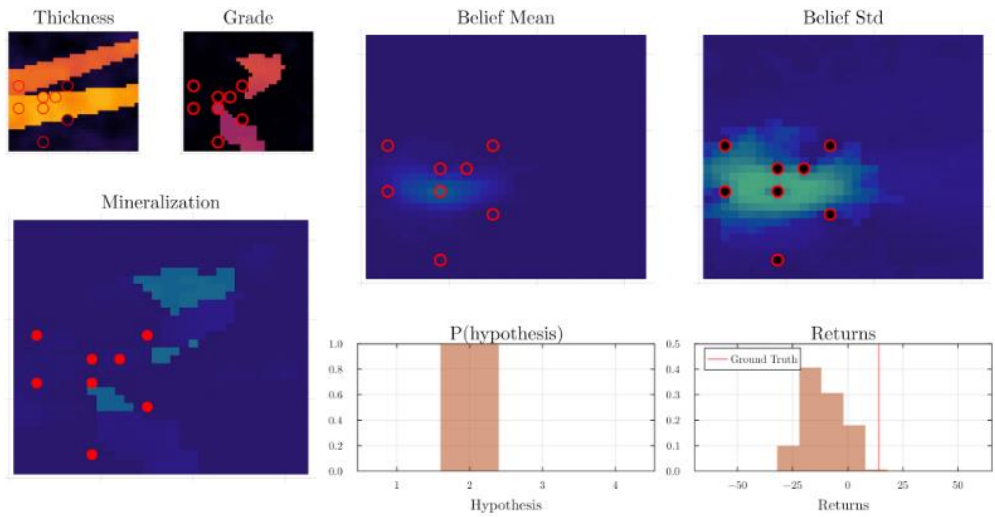

Figure 10: when planning under the wrong model hypothesis, significant portion of the mineralization is not identified in this case.

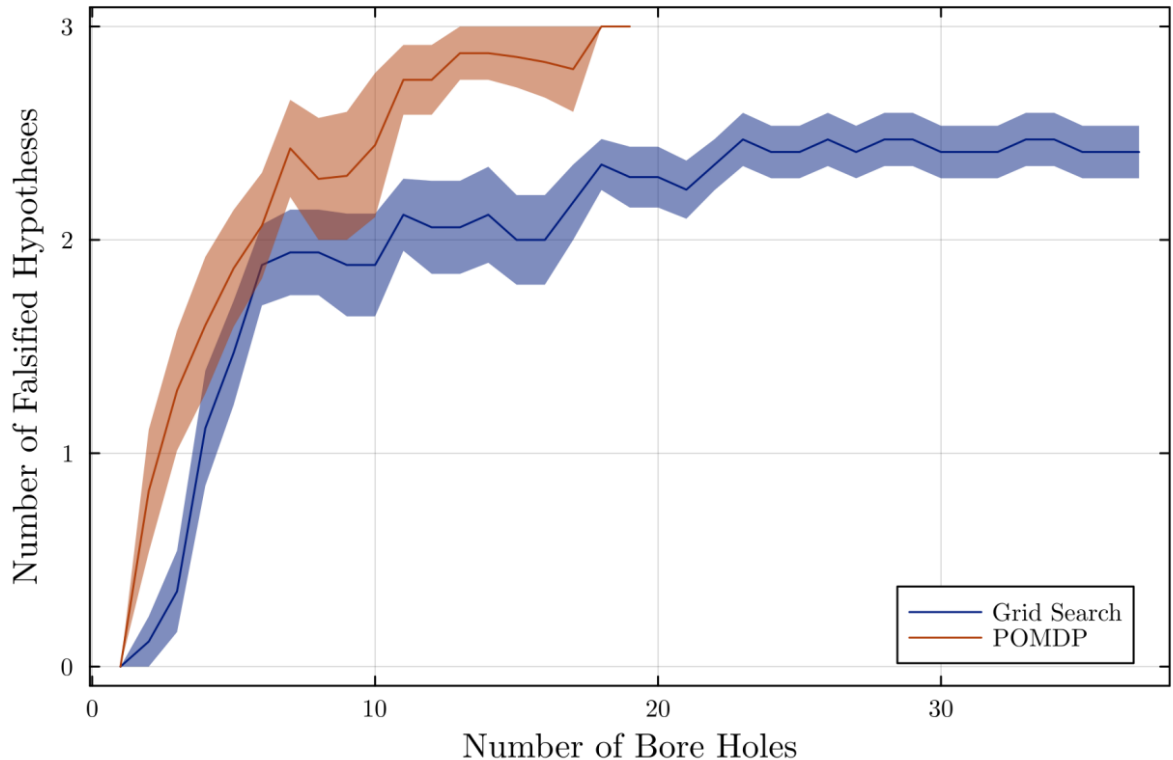

Figure 11: comparison of the number of bore holes (averaged over 17 trials) at which all hypotheses are falsified for both the POMDP agent and the grid baseline

## 5. Discussion

We presented a completely new methodology for drill-planning in mineral exploration and appraisal. Here we discuss further some elements, both non-scientific and scientific, related to its implementation in actual cases. The reality of today's mineral exploration is that such endeavours are mostly performed by so-called junior mining companies. These companies work with small drilling budgets, and additional funding is provided by investors when drilling shows encouraging results. This means that the "reward" for drilling is not the reduction of uncertainty on geological hypothesis, but targeting zones deemed high in mineralization. What we propose here is completely different, and also requires a very different investment strategy. We have shown that in this example, reducing uncertainty in geological hypotheses is critical to minimizing the number of boreholes in the long run. Instead of drilling one borehole at a time, we plan the first borehole knowing that a possible long sequence will be drilled, each with different outcomes. POMDP models plans with the idea that future information will reveal something about the state of the subsurface, which can help the present decisions.

This very idea also suggests that there are various ways of formulating the reward. In our case we used a very simple economic model, since the focus of this paper is methodology and not actual cases with real economic implications. However, in real cases one may need to carefully decide on the reward function itself. A reward function may focus on uncertainty reduction on grade-tonnage curves directly. One may also decide to add an uncertainty reward to the total reward. For example, one may choose to include reduction of uncertainty of a geological hypothesis, such reward would enter as a form of entropy that measures such uncertainty. Decision science does not tell us what reward should be included, hence understanding this subjectivity is certainly a case for future investigation.

## 6. Conclusions

Accelerating discovery of critical mineral will require efficient data acquisition by focusing on reducing uncertainty of properties of interest such as geological hypothesis, grades and tonnes. In this paper, we present an entirely new approach to exploration, in particular to drill planning. At early-stage exploration, the leading uncertainty is usually the definition of conceptual geological hypotheses. These hypotheses are used to create 3D geological models whose uncertainty is dominated by what hypotheses is chosen. In a real exploration setting it is more likely than not that human generated hypotheses early at the exploration stage will eventually be proven incorrect, i.e. falsified, after drilling proceeds. In this paper we develop an artificial intelligent agent that can detect this falsification early on, and hence prevent drilling under completely incorrect hypothesis. We have shown that are agent in the examples presented, is much more efficient that the industry standard of grid-based drilling, and that its first implementation has led to the fast appraisal of an ultra high-grade deposit in Zambia.

## 7. Appendix

Note that due to the synthetic nature of the geological models, the following parameters are unitless and should not be interpreted as representing any particular unit. The 2D grids are each 32 grid cells in each dimension. The Gaussian process models used for grade and thickness all use the same Matern kernel with marginal standard deviation of 0.1 and a correlation length of 3 grid cells. The mean thickness outside of a graben was 1.0 and inside a graben was 7.5. The mean grade outside of an altered geochemical domain was 0.0 and 0.085 inside a geochemical domain. The noise on both thickness and grade measurements was gaussian with a mean of 0.0 and a standard deviation of 0.001. Each side of the graben is defined by a bottom location (Normal with $mean = 11, std = 6$) and a width (normal distribution with $mean = 6, std = 2$). Each geochemical domain is defined by a polygonal area with a center (normal with $mean = 16$ and $std = 8$ in each dimension) and 10 evenly spaced points around it that each have a distance from the center that is normally distributed with a $mean = 5$ and $std = 2.5$. The null hypothesis was selected to be a mixture of non-spatially correlated Gaussians for both the thickness and grade.

**Code/data availability**

https://github.com/ancorso/HierarchicalMineralExploration.jl

**Author contribution**

John Mern: conceptualization, coding, analysis; Anthony Corso: conceptualization, coding, analysis; Damian Burch (analysis; review; editing); Kurt House (conceptualization, review, editing), Jef Caers (conceptualization, writing, analysis)

**Competing interests**

The authors declare that they have no conflict of interest.

# References

Abedi, M., Norouzi, G.H. and Torabi, S.A., 2013. Clustering of mineral prospectivity area as an unsupervised classification approach to explore copper deposit. *Arabian Journal of Geosciences*, *6*, pp.3601-3613.

Bickel, J.E., Smith, J.E. and Meyer, J.L.: Modeling dependence among geologic risks in sequential exploration decisions. SPE Reservoir Evaluation & Engineering, 11(02), pp.352-361. https://doi.org/10.2118/102369-PA, 2008.

Brechtel, S., Gindele, T. and Dillmann, R.: Probabilistic decision-making under uncertainty for autonomous driving using continuous POMDPs. In 17th international IEEE conference on intelligent transportation systems (ITSC) (pp. 392-399). IEEE, 2014.

Caers, J., 2018. Bayesianism in the Geosciences. *Handbook of Mathematical Geosciences: Fifty Years of IAMG*, pp.527-566.

Caers, J., Scheidt, C., Yin, Z., Mukerji, T. and House, K:. Efficacy of Information in Mineral Exploration Drilling. Nat Resour Res 31, 1157–1173 https://doi-org.stanford.idm.oclc.org/10.1007/s11053-022-10030-1, 2022.

Chaslot, G., Bakkes, S., Szita, I. and Spronck, P.: Monte-carlo tree search: A new framework for game ai. In Proceedings of the AAAI Conference on Artificial Intelligence and Interactive Digital Entertainment (Vol. 4, No. 1, pp. 216-217), 2008.

Clare, A.P. and Cohen, D.R., 2001. A comparison of unsupervised neural networks and k-means clustering in the analysis of multi-element stream sediment data. *Geochemistry: Exploration, Environment, Analysis*, *1*(2), pp.119-134.

Corso, A, (2024) ancorso/HierarchicalMineralExploration.jl: Intelligent Prospector v2.0, Zenodo. DOI: 10.5281/zenodo.13850925

Demsey, H. "Bill Gates-backed mining company discovers vast Zambian copper deposit" Financial Times, February 4, 2024.

Diggle P. and Lophaven S.: Bayesian geostatistical design. Scandinavian Journal of Statistics, 33(1): 53–64. https://doi.org/10.1111/j.1467-9469.2005.00469.x, 2006.

Dumakor-Dupey, N.K. and Arya, S., 2021. Machine learning—a review of applications in mineral resource estimation. *Energies*, *14*(14), p.4079.
29

Eidsvik, J. and Ellefmo, S.L.: The value of information in mineral exploration within a multi-Gaussian framework, Mathematical Geosciences, 45, 777-798. https://doi.org/10.1007/s11004-013-9457-2, 2013.

Eidsvik, J., Martinelli, G., Bhattacharjya, D.: Sequential information gathering schemes for spatial risk and decision analysis applications. Stoch. Environ. Res. Risk Assess. 32 (4), 1163–1177. https://doi.org/10.1007/s00477-017-1476-y, 2018.

Emery, X., Hernández, J., Corvalán, P., and Montaner, D.: Developing a cost-effective sampling design for forest inventory, in Ortiz, J. M., and Emery, X., eds., Proceedings of the Eighth International Geostatistics Congress, Vol. 2: Gecamin, p. 1001–1010, 2008.

Gazley, M.F., Collins, K.S., Roberston, J., Hines, B.R., Fisher, L.A. and McFarlane, A., 2015. Application of principal component analysis and cluster analysis to mineral exploration and mine geology. In *AusIMM New Zealand branch annual conference* (Vol. 2015, pp. 131-139). Dunedin New Zealand.

Grema, Alhaji Shehu & Cao, Yi.: Optimization of petroleum reservoir waterflooding using receding horizon approach. Proceedings of the 2013 IEEE 8th Conference on Industrial Electronics and Applications, ICIEA 2013. 397-402. 10.1109/ICIEA.2013.6566402, 2013.

Grigorescu, S., Trasnea, B., Cocias, T. and Macesanu, G.: A survey of deep learning techniques for autonomous driving. Journal of Field Robotics, 37(3), pp.362-386. https://doi-org.stanford.idm.oclc.org/10.1002/rob.21918, 2020.

Hall, T., Scheidt C., Wang L., Zhen, Y, Mukerji, T., Caers, J.: Sequential value of information for subsurface exploration drilling, Natural Resources Research, 2022 (accepted for publication).

Heuvelink G.B.M., Brus D., and de Gruijter J.J.: Optimization of sample configurations for digital mapping of soil properties with universal kriging. In: Lagacherie P, McBratney A, Voltz M, editors. Digital soil mapping: an introductory perspective. Amsterdam, The Netherlands: Elsevier; p. 139–153 https://doi.org/10.1016/s0166-2481(06)31011-2, 2006.

Hitzman, M., Kirkham, R., Broughton, D., Thorson, J. and Selley, D., 2005. The sediment-hosted stratiform copper ore system.

Jung, D. and Choi, Y., 2021. Systematic review of machine learning applications in mining: Exploration, exploitation, and reclamation. *Minerals*, *11*(2), p.148.

Kochenderfer, M.J., Wheeler, T.A. and Wray, K.H., 2022. *Algorithms for decision making*. MIT press.